\PassOptionsToPackage{table}{xcolor}

\documentclass[sigconf,authordraft,review=false]{acmart}
\makeatletter
\def\@author@block@height{5.8\baselineskip}  
\makeatother

\usepackage{multirow}
\usepackage{graphicx}
\usepackage{booktabs}
\usepackage{algorithm}
\usepackage{algorithmic}
\usepackage{multirow}

\AtBeginDocument{%
  }

\setcopyright{acmlicensed}
\copyrightyear{2018}
\acmYear{2018}
\acmDOI{XXXXXXX.XXXXXXX}
\acmConference[Conference acronym 'XX]{Make sure to enter the correct
  conference title from your rights confirmation email}{June 03--05,
  2018}{Woodstock, NY}
\acmISBN{978-1-4503-XXXX-X/2018/06}
\renewcommand\footnotetextcopyrightpermission[1]{}

\begin{document}
\title{LG-HCC: Local Geometry-Aware Hierarchical Context Compression for 3D Gaussian Splatting}

\author{Xuan Deng}
\affiliation{%
  \institution{Harbin Institute of Technology, Peng Cheng Laboratory}
  \city{ Shenzhen}
  \country{China}}
\email{dengxuan168168@gmail.com}

\author{Xiandong Meng}
\affiliation{%
  \institution{Peng Cheng Laboratory}
  \city{ Shenzhen}
  \country{China}}
\email{mengxd@pcl.ac.cn}

\author{Hengyu Man}
\affiliation{%
  \institution{Harbin Institute of Technology}
  \city{Harbin}
  \country{China}}
\email{manhengyu@hotmail.com}

\author{Qiang Zhu}
\affiliation{%
  \institution{Peng Cheng Laboratory}
  \city{ Shenzhen}
  \country{China}}
\email{zhuqiang@std.uestc.edu.cn}

\author{Tiange Zhang}
\affiliation{%
  \institution{Peng Cheng Laboratory}
  \city{ Shenzhen}
  \country{China}}
\email{zhangtg.@pcl.ac.cn}

\author{Debin Zhao}
\affiliation{%
  \institution{Harbin Institute of Technology}
  \city{Harbin}
  \country{China}}
\email{dbzhao@hit.edu.cn}

\author{Xiaopeng Fan}
\affiliation{%
  \institution{Harbin Institute of Technology, Peng Cheng Laboratory, Harbin Institute of Technology Suzhou Research Institute}
  \city{Harbin}
  \country{China}}
\email{fxp@hit.edu.cn}







\renewcommand{\shortauthors}{Trovato et al.}



\begin{abstract}
Although 3D Gaussian Splatting (3DGS) enables high-fidelity real-time rendering, its prohibitive storage overhead severely hinders practical deployment. Recent anchor-based 3DGS compression schemes reduce gaussian redundancy through some advanced context models. However, they overlook explicit geometric dependencies, leading to structural degradation and suboptimal rate-distortion performance.
In this paper, we propose a \textbf{L}ocal \textbf{G}eometry-aware \textbf{H}ierarchical \textbf{C}ontext \textbf{C}ompression framework for 3DGS  (\textbf{LG-HCC}) that incorporates inter-anchor geometric correlations into anchor pruning and entropy coding for compact representation. Specifically, we introduce an Neighborhood-Aware Anchor Pruning (NAAP) strategy, which evaluates anchor importance via weighted neighborhood feature aggregation and then merges low-contribution anchors into salient neighbors, yielding a compact yet geometry-consistent anchor set. Moreover, we further develop a hierarchical entropy coding scheme, in which coarse-to-fine priors are exploited through a lightweight Geometry-Guided Convolution (GG-Conv) operator to enable spatially adaptive context modeling and rate-distortion optimization. Extensive experiments show that LG-HCC effectively alleviates structural preservation issues, achieving superior geometric integrity and rendering fidelity while reducing storage by up to 30.85$\times$ compared to the Scaffold-GS baseline on the Mip-NeRF360 dataset.

\end{abstract}



\maketitle

\begin{figure}[t]
\centering
\includegraphics[height=0.6\linewidth]{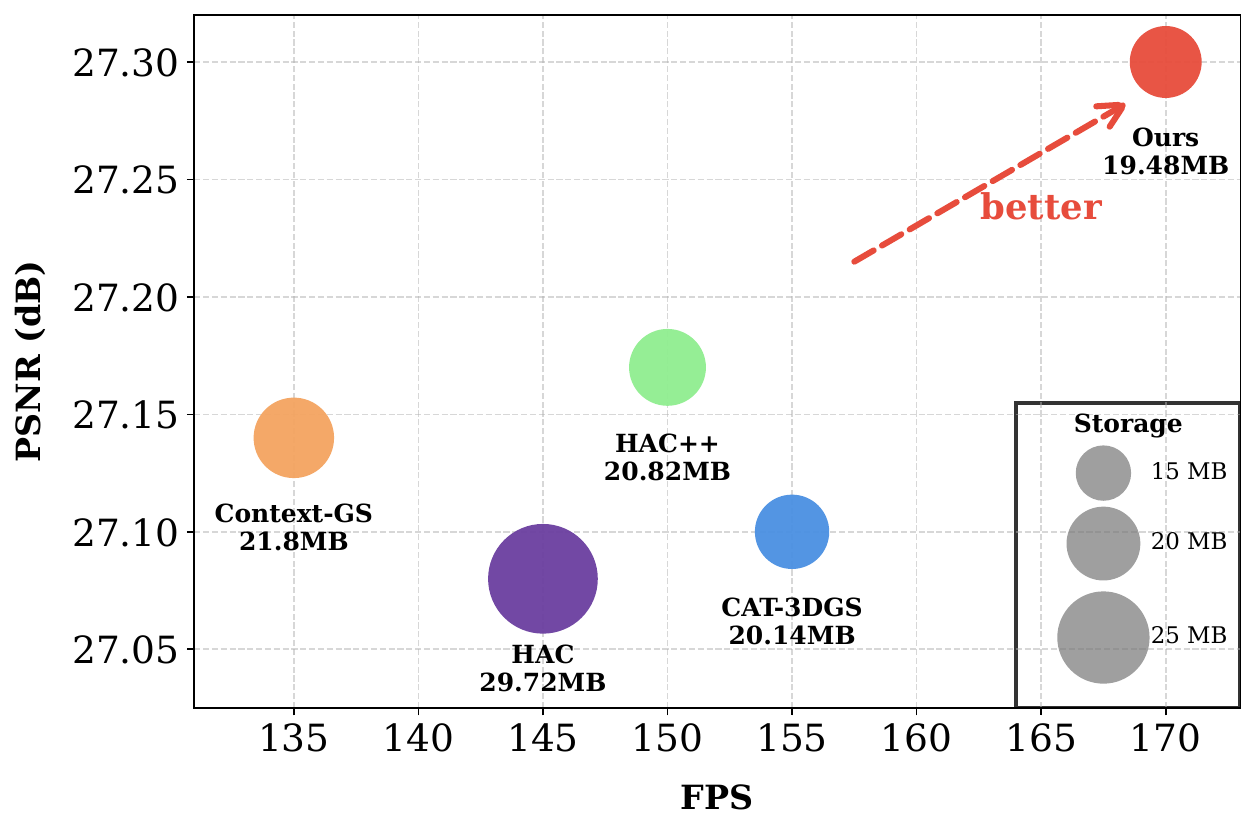} 
\caption{Trade-off between fidelity (PSNR) and rendering speed (FPS) for different methods on the BungeeNeRF dataset~\cite{xiangli2022bungeenerf}. Circle size denotes storage cost, where smaller circles correspond to lower storage.}
\label{fig:fps}
\end{figure}

\section{Introduction}

3D Gaussian Splatting (3DGS)~\cite{kerbl20233d} has emerged as a powerful explicit 3D representation that models scenes with a set of anisotropic Gaussians with learnable attributes, including position, covariance, opacity, and spherical harmonics, enabling efficient optimization and high-quality novel-view synthesis. Its optimized rasterization supports fast convergence and real-time rendering, offering a strong alternative to implicit neural representations. However, high-quality reconstruction typically requires millions of primitives, leading to considerable storage and memory overhead. This redundancy hinders practical deployment, especially on resource-constrained platforms, and has motivated extensive research on compact 3DGS representations and compression.

Early 3DGS compression methods~\cite{lee2024compact,niedermayr2024compressed,girish2024eagles,fan2024lightgaussian,fang2024mini,mallick2024taming,kim2024color} mainly reduce storage through primitive pruning or vector quantization. While effective, they focus on compressing attribute value of each individual gaussian and largely ignore the inherent geometric correlation among Gaussians. Although such dependencies have been widely exploited in image and video compression~\cite{cheng2020learned,he2021checkerboard,li2024neural,li2023neural,sheng2022temporal,jia2025towards}, modeling them within 3DGS remains non-trivial due to the irregularly distribution of Gaussian primitives in 3D space. Locality-aware-GS~\cite{shin2025locality} attempts to incorporate local geometric relations directly over massive unstructured Gaussian primitives, but the cost of neighborhood correlation search becomes prohibitive at large scale. Conversely, Anchor-based representations, e.g., Scaffold-GS~\cite{lu2024scaffold}, alleviate this issue by hierarchically clustering primitives around anchors and thus provide a more compact and structured representation. Nevertheless, Scaffold-GS treats anchors as isolated entities, leaving the local geometric correlations among neighboring anchors underexplored.

To further alleviate spatial redundancy, recent advances have incorporated more structured priors into anchor-based compression framework. For example, inspired by NeRF feature grids~\cite{mildenhall2021nerf,muller2022instant}, HAC~\cite{chen2024hac} and HAC++~\cite{chen2025hac++} leverage hash grids for context modeling.
Context-GS~\cite{wang2024ContextGS} takes a different approach by organizing anchors hierarchically to capture dependencies across representation levels,  thus improving entropy modeling.
Despite these advances, a fundamental limitation persists: current methods largely overlook the intrinsic geometric structures among anchors. Grid-based context models (e.g. HAC) rely on coarse extrinsic discretization and often fail to capture fine-grained geometric relationships in irregular anchor clouds. Hierarchical methods such as Context-GS~\cite{wang2024ContextGS}, despite their stronger representational capacity, typically construct multi-level context structures on top of heuristic pruning strategies, such as opacity-based anchor removal. By directly discarding low-importance anchors, these methods may disrupt local geometric continuity and weaken the structural foundation required for reliable coarse-to-fine context prediction.  Since finer levels can no longer inherit sufficiently stable geometric priors from the base layers, ultimately leading to suboptimal structure preservation and entropy estimation.

In this work, we argue that exploiting spatial context is fundamental to maximizing the compression efficiency of 3DGS. Instead of treating anchors as isolated containers of attributes, we represent them as interconnected nodes within a geometric graph. This perspective allows us to capture strong spatial corrleations within local neighborhoods. Based on this insight, we propose a \textbf{L}ocal \textbf{G}eometry-aware \textbf{H}ierarchical \textbf{C}ontext \textbf{C}ompression framework for 3D Gaussian spaltting (\textbf{LG-HCC}) that incorporates inter-anchor geometric correlations into anchor pruning and entropy coding for compact representation. Specifically, we first introduce Neighborhood-Aware Anchor Pruning (NAAP), which evaluates anchor importance via graph aggregation and then merges low-contribution anchors into salient neighbors. Building on the preserved local structure, we further design a hierarchical entropy coding scheme with geometry-guided context. Specifically, we design a lightweight  Geometry-Guided Convolution (GG-Conv) operator to extract contextual cues from local geometric configurations. This allows our entropy model to learn local geometry-aware context priors; these priors are spatially adaptive to varying levels of detail and geometrically consistent with the anchor distribution, leading to more accurate probability estimation and reduced bit-rates. As shown in Figure~\ref{fig:fps}, experiments on the BungeeNeRF~\cite{xiangli2022bungeenerf}  dataset show that LG-HCC achieves a PSNR-FPS trade-off than existing methods while requiring lower storage cost.
Our main contributions are summarized as follows:
\begin{itemize}
    \item We propose LG-HCC, a local geometry-aware hierarchical context compression framework for 3D Gaussian Splatting that reformulates anchor compression via geometry-induced graphs for consistent pruning and efficient hierarchical entropy modeling.
    

    \item To preserve the local geometric perception capability of anchors, we propose Neighborhood-Aware Anchor Pruning (NAAP). NAAP is a graph-based strategy that adaptively prunes less significant anchors by evaluating their importance within the local geometric context, thus achieving an optimal sparse representation.
    
    \item To exploit local geometric correlation in sparsified Gaussians, we propose hierarchical geometry-guided context modeling via a lightweight Geometry-Guided Convolution (GG-Conv). By capturing spatially adaptive priors over irregular neighborhoods, our method achieves accurate probability estimation and state-of-the-art compression with high rendering quality.
\end{itemize}

\section{Related Work}

\subsection{Graph-based 3D data modeling}

Graph Signal Processing (GSP)~\cite{ortega2018graph,yang2024spectrally} provides a framework for analyzing signals on irregular domains by modeling data as nodes in a graph. In 3D vision, it has been widely applied to point cloud analysis (e.g., DGCNN~\cite{wang2019dynamic} and PointNet++~\cite{qi2017pointnet,qi2017pointnet++}), where local graphs encode geometric relations for feature aggregation. Similar principles have also been applied to point cloud compression, where exploiting inter-point correlations improves coding efficiency~\cite{fan2022d,deng2025pvinet,gao2026overview}.
In this paper, we propose LG-HCC, which reformulates anchor-based 3DGS as an irregular geometric graph to jointly optimize anchor pruning and entropy coding via local geometry guidance, achieving significant bitrate reduction while preserving high-fidelity structural details.

\subsection{Compact 3DGS Representation}
3D Gaussian Splatting (3DGS)~\cite{kerbl20233d} achieves high-fidelity and efficient radiance field rendering, yet its explicit and unstructured Gaussian representation is less regular than the grid-based representations of NeRFs~\cite{mildenhall2021nerf,bagdasarian24073dgs,wu2024recent}, leading to prohibitive storage overhead that necessitates advanced compression techniques.

Existing methods primarily tackle the storage challenge of 3D Gaussian Splatting by either reducing the number of primitives or learning compact attribute representations, without explicit rate-distortion optimization. For example, pruning-based approaches~\cite{lee2024compact,fan2024lightgaussian,navaneet2023compact3d,papantonakis2024reducing,ali2024trimming,ren2024octree,cheng2024gaussianpro,liu2024atomgs} eliminate less contributive Gaussians using heuristic criteria, such as learnable masks, gradient magnitudes, or view-dependent significance. Complementary attribute compression techniques include spherical harmonics coefficient pruning~\cite{morgenstern2024compact} and vector quantization~\cite{wang2024end,lee2024compact,girish2024eagles}.
In addition to these primitive-level techniques, a representative anchor-based approach is Scaffold-GS~\cite{lu2024scaffold}, which uses anchor points to hierarchically distribute local 3D Gaussians and predicts their attributes on-the-fly conditioned on viewing direction and distance within the view frustum. 
Inspired by the anchor-based representation of Scaffold-GS~\cite{lu2024scaffold}, a series of recent works has focused on rate-distortion (RD) optimized compression by improving entropy modeling through the exploitation of spatial or structural priors. These anchor-based methods have achieved impressive storage reduction while maintaining high rendering quality. For instance, HAC~\cite{chen2024hac} and HAC++~\cite{chen2025hac++} employ hash-grid-based spatial organization to derive compact context representations. Context-GS~\cite{wang2024ContextGS} and CompGS~\cite{liu2024compgs} further exploit hierarchical structure and anchor-primitive dependencies to improve entropy coding. Meanwhile, CAT-3DGS~\cite{zhan2025cat} adopts channel-wise autoregressive models to capture intra-attribute correlations, and MoP~\cite{liu20253d} enhance density estimation through a Mixture-of-Priors formulation.

While these methods achieve impressive rate-distortion performance in 3D Gaussian Splatting compression, we argue that the core of effective compression lies in the entropy model. A well-designed entropy model can leverage local geometric correlations among anchors, enabling efficient coding and reducing storage requirements. However, current anchor-based 3DGS compression methods still have room for improvement in entropy model design, particularly in exploiting local geometric priors. Drawing inspiration from geometry-aware modeling in point cloud compression and graph neural networks, we propose a local geometry-aware entropy model. This model represents the unstructured anchor cloud as an irregular geometric graph and uses a lightweight Geometry-Guided Convolution (GG-Conv) to adaptively aggregate geometry-aware contextual priors from neighboring anchors.

\section{Methodology}
\subsection{Preliminaries}

\textbf{3D Gaussian Splatting (3DGS).}
3DGS~\cite{kerbl20233d} utilizes a collection of anisotropic 3D neural Gaussians to depict the scene so that the scene can be efficiently rendered using a tile-based rasterization technique. Beginning from a set of Structure-from-Motion (SfM) points, each Gaussian point is represented as follows:
\begin{equation}
    G(\mathbf{p}) = \exp\left(-\frac{1}{2}(\mathbf{p}-\boldsymbol{\mu})^\top \boldsymbol{\Sigma}^{-1}(\mathbf{p}-\boldsymbol{\mu})\right),
\end{equation}
where $\mathbf{p}$ denotes the coordinates in the 3D scene, and $\boldsymbol{\mu}$ and $\boldsymbol{\Sigma}$ represent the mean position and covariance matrix of the Gaussian point, respectively. To ensure the positive semi-definiteness of $\boldsymbol{\Sigma}$, it is parameterized as 
$\boldsymbol{\Sigma} = \mathbf{R} \mathbf{S} \mathbf{S}^\top \mathbf{R}^\top$, 
where $\mathbf{R}$ and $\mathbf{S}$ denote the rotation and scaling matrices, respectively. 
Furthermore, each neural Gaussian possesses an opacity attribute $\alpha \in \mathbb{R}^1$ 
and view-dependent color $\mathbf{c} \in \mathbb{R}^3$, modeled via spherical harmonics~\cite{zhang2022differentiable}. 
All attributes of the neural Gaussians, i.e., $\{\boldsymbol{\mu}, \mathbf{R}, \mathbf{S}, \alpha, \mathbf{c}\}$, 
are learnable and optimized by minimizing the reconstruction loss of images rendered through tile-based rasterization.


\begin{figure*}[!t]
\centering
\includegraphics[width=1\textwidth]{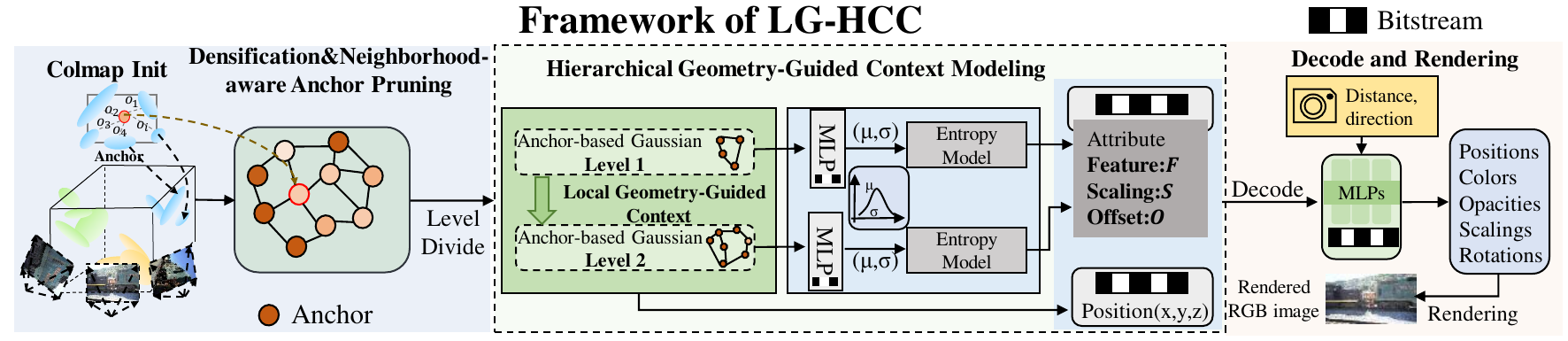} 
\caption{\textbf{Overview of the LG-HCC framework.} 
Following Colmap initialization and densification, \textbf{Neighborhood-Aware Anchor Pruning (NAAP)} adaptively merges low-contribution anchors into salient neighbors via a local graph to ensure geometric consistency. 
The preserved anchors then undergo \textbf{Hierarchical Geometry-Guided Context Modeling} for efficient entropy coding. 
Finally, decoded attributes generate 3D Gaussians via learnable offsets for high-fidelity rasterization.}
\label{fig:framework}
\end{figure*}

\subsection{Overview}
\label{subsec:overview}

As illustrated in Figure~\ref{fig:framework}, LG-HCC builds upon the anchor-based representation of Scaffold-GS~\cite{lu2024scaffold}. We define each anchor as a tuple $\mathbf{a} = \{\mathbf{p}, \mathbf{f}, \mathbf{s}, \mathbf{o}\}$, comprising position $\mathbf{p} \in \mathbb{R}^3$, feature $\mathbf{f} \in \mathbb{R}^{C}$, scaling factor $\mathbf{s} \in \mathbb{R}^3$, and offsets $\mathbf{o} \in \mathbb{R}^{K \times 3}$. The pipeline begins with anchor densification, followed by Neighborhood-Aware Anchor Pruning (NAAP) module (Sec.~\ref{sec:naap}), which sparsifies the representation by merging redundant anchors based on a local geometric graph. The remaining anchors are then structured into a multi-level hierarchy to enable hierarchical geometry-guided context modeling (Sec.~\ref{subsec:entropy_coding}). Specifically, a lightweight Geometry-Guided Convolution (GG-Conv) aggregates neighborhood features to derive local geometry-aware context priors for entropy coding. Finally, 3D Gaussians are reconstructed from the decoded anchors via learnable offsets for rasterization.


\begin{figure}[!t]
\centering
\includegraphics[height=0.8\linewidth]{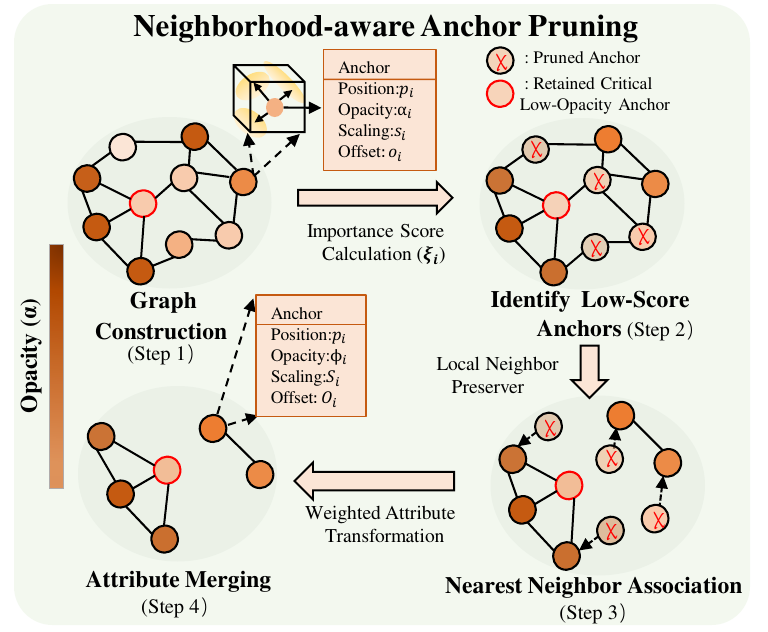} 

\caption{Neighborhood-Aware Anchor Pruning (NAAP) pipeline starts by constructing an anchor-based geometric graph, with nodes colored by average opacity (darker = higher). Importance scores ($\xi_i$) are computed via neighborhood aggregation. Low-score anchors are identified as redundant and associated with their nearest salient neighbor (Local Neighbor Preserver). Instead of direct removal, a Weighted Attribute Transformation merges attributes from pruned anchors (opacity $\bar\alpha_i$, scaling $s_i$, offset $o_i$) into the nearest retained anchor, yielding fused attributes ($\phi_i$, $O_i$, $S_i$, etc.) to preserve geometric and appearance in a compact structure.} 


\label{fig:pruning}
\end{figure}

\subsection{Neighborhood-Aware Anchor Pruning}
\label{sec:naap}

Conventional 3DGS pruning strategies typically rely on point-wise attribute thresholds (e.g., opacity) to remove less important Gaussian anchors. However, such independent evaluation ignores local geometric distributions, may inadvertently eliminate structurally  essential anchors, especially in low-density regions.
To address this limitation, we propose Neighborhood-Aware Anchor Pruning (NAAP), which evaluates anchor importance within a local geometric neighborhood context and preserves structural continuity by adaptively aggregating redundant features into their local neighborhood.
As illustrated in Figure~\ref{fig:pruning}, the proposed NAAP consists of four steps.

\noindent \textbf{Step 1 \& 2: Graph Construction and Importance Evaluation.}
We initially construct a local geometric graph $\mathcal{G} = (\mathcal{A}, \mathcal{E})$ to capture the local geometric correlations among anchors, where $\mathcal{A} = \{\mathbf{a}_1, \dots, \mathbf{a}_z\}$ represents the set of anchors and $  \mathcal{E}  $ denotes the edges connecting neighboring anchors, each anchor $\mathbf{a}_i$ possesses a position $\mathbf{p}_i$ and an accumulated average opacity $\bar{\alpha}_i$, which is derived by accumulating the opacity values of its associated neural Gaussians over 
$\mathcal{I}$ training iterations. We connect each anchor to its $K$-nearest neighbors within a radius $r$, forming the local neighborhood $\mathcal{N}(i)$, which provides a lightweight representation of local geometric structure for importance estimation.

To distinguish structurally essential anchors from noise or redundant anchors, we define a neighborhood-weighted importance score.
Specifically, for each anchor (taking $\mathbf{a}_i$ as an example), we first compute a distance-based weight  $w_{ij} = (\|\mathbf{p}_i - \mathbf{p}_j\|_2 + \epsilon)^{-1}$ between between itself and its neighbor $\mathbf{a}_j \in \mathcal{N}(i)$, which assigns larger weights to closer neighbors.
Then, a smoothed neighborhood opacity $\Phi_i$ is  calculated to aggregate local context:
\begin{equation}
    \Phi_i = \frac{\bar{\alpha}_i + \sum_{j \in \mathcal{N}(i)} w_{ij} \bar{\alpha}_j}{1 + \sum_{j \in \mathcal{N}(i)} w_{ij}}.
\end{equation}
The final importance score $\xi_i$  is formulated as a linear interpolation between the anchor-specific opacity and its neighborhood-aware observation:
\begin{equation}
    \xi_i = (1 - \lambda)\bar{\alpha}_i + \lambda \Phi_i,
\end{equation}
where $\lambda \in [0, 1]$ is a balancing factor.
Given a threshold $\tau$, we identify redundant anchors using the pruning mask: 
\begin{equation}
    \mathcal{M}_{prune} = \{\mathbf{a}_i \mid \xi_i < \tau\}.
\end{equation}
This design enables the anchors residing in structurally significant clusters to be preserved even if their individual opacity is relatively low.

\noindent \textbf{Step 3 \& 4: Neighbor Association and Attribute Merging.}
Unlike previous pruning strategies that directly discard anchors in $\mathcal{M}_{\text{prune}}$, we instead consolidate their information into the surviving geometry via a dedicated \textbf{Weighted Attribute Transformation}. 
For each redundant anchor $\mathbf{a}_r \in \mathcal{M}_{\text{prune}}$, we identify its nearest \textit{salient neighbor} $\mathbf{a}_k \in \mathcal{A} \setminus \mathcal{M}_{\text{prune}}$ as the target for information transfer. 
Let lowercase $\theta \in \{\mathbf{o}, \mathbf{s}, \alpha\}$ and uppercase $\Theta \in \{\mathbf{O}, \mathbf{S}, \phi\}$ denote the raw and fused attributes (offsets, scaling, and opacity), respectively. 
The transformation is governed by:
\begin{equation}
    \Theta_k = (1 - \gamma) \theta_k + \gamma \theta_r,
    \label{eq:weighted_transformer}
\end{equation}
where $\gamma$ controls the contribution of the pruned anchor. 
This integration effectively condenses the geometric and appearance attributes of removed regions into their nearest neighbors ($\theta \to \Theta$), ensuring that comprehensive scene information is preserved rather than lost. 
Finally, eliminating the anchors in $\mathcal{M}_{\text{prune}}$ yields a compact, locally geometry-consistent representation.


\begin{figure*}[!t]
\centering
\includegraphics[width=1\textwidth]{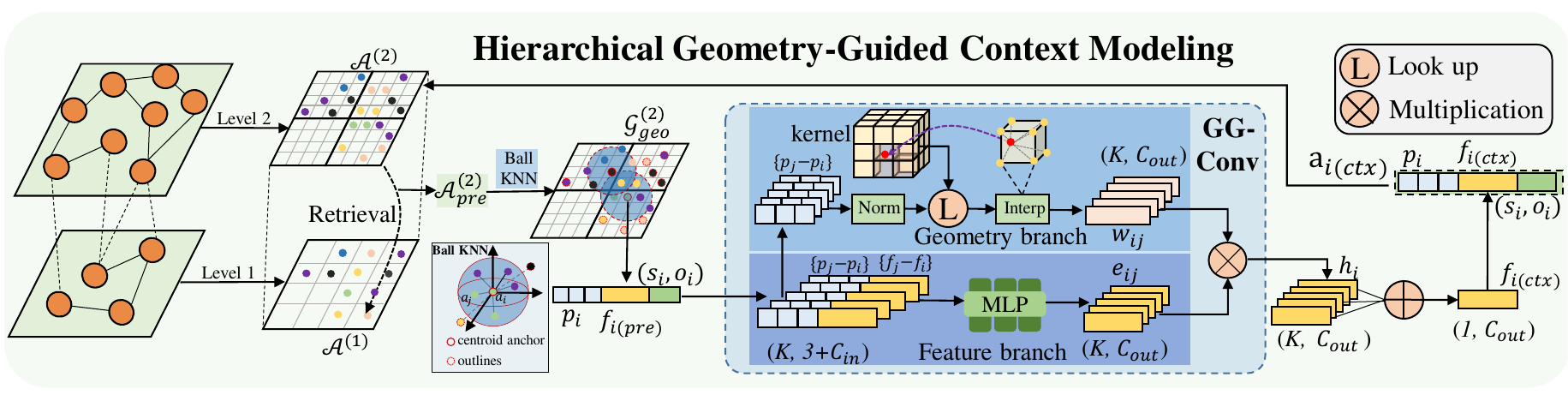} 
\caption{
This module exploits cross-level correlations by mapping Level 1 attributes to Level 2 query anchors via a deterministic mapping $\mathcal{M}$ to obtain the preliminary attributes $\mathcal{A}_{pre}^{(2)}$. Then, a $k$-NN graph is constructed over $\mathcal{A}_{pre}^{(2)}$ to form the graph-based geometric prior $\mathcal{G}_{geo}^{(2)}$. Within the GG-Conv (dashed box):
\textbf{(1) Geometry branch:} Relative offsets $\Delta \mathbf{p}_{ij}$ query a learnable 3D weight table via trilinear interpolation to generate dynamic weights $\mathbf{w}_{ij}$ encoding the spatial layout.
\textbf{(2) Feature branch:} Residual features $\Delta \mathbf{f}_{ij}^{(1)}$ and offsets $\Delta \mathbf{p}_{ij}$ are transformed by MLP $\phi$ into embeddings $\mathbf{e}_{ij}$.
Finally, $\mathbf{w}_{ij}$ modulates $\mathbf{e}_{ij}$ to produce the local geometry-aware feature $\mathbf{h}_i$, which serves as the final context $\mathbf{a}_{i(ctx)}$ for Level 2 entropy coding.}
\label{fig:context_module}
\end{figure*}


As shown in Figure~\ref{fig:pruning}, conventional threshold-based pruning would discard low-opacity anchors (highlighted by the red circle) induced by optimization noise.
In contrast, NAAP can identify non-salient nodes within high-density clusters, which often play a crucial role in modeling key scene details during rendering.



\subsection{Hierarchical Geometry-Guided Context Modeling}
\label{subsec:entropy_coding}
While hierarchical context models~\cite{wang2024ContextGS,liu2024compgs,liu2024hemgs} have shown promising performance for anchor-based 3DGS compression, they often suffer from structural inefficiencies. Specifically, existing approaches typically rely on heuristic neighbor pooling or rigid spatial partitioning, which may fail to preserve precise geometric correspondence across hierarchy levels. As a result, the retrieved context is not always well-aligned with the query anchors, leading to suboptimal entropy estimation.
To overcome these limitations, we propose a streamlined two-level autoregressive model that explicitly enforces geometric alignment between coarse and fine anchors. As shown in Figure~\ref{fig:context_module}, by strictly aligning the coarse geometry (Level 1) with fine attributes (Level 2), our method achieves precise context retrieval and refinement via a following three-step process.

\vspace{5pt}
\noindent \textbf{1) Hierarchical Partitioning.} To leverage multi-scale spatial correlations, we organize the anchor set $\mathcal{V}$ into a two-level hierarchy ($\mathcal{L}=2$) following Context-GS \cite{wang2024ContextGS}. We first derive a coarse representation $\mathcal{A}^{(1)}$ by quantizing anchor positions with a scaled voxel size $\epsilon_1 = s \cdot \epsilon_0$:
\begin{equation}
\mathcal{A}^{(1)} = \{\mathbf{a}_j \in \mathcal{V} \mid j = \min \{i : \lfloor p_i / \epsilon_1 \rfloor \}\},
\label{6}
\end{equation}
where $\epsilon_0$ is the initial fine-grained voxel size and $\epsilon_1$ represents the coarse-grained resolution controlled by the scaling factor $s$. The fine level is then defined as $\mathcal{A}^{(2)} = \mathcal{V} \setminus \mathcal{A}^{(1)}$ to ensure a disjoint structure. This hierarchy allows $\mathcal{A}^{(1)}$ to serve as a coarse spatial prior for the entropy coding of $\mathcal{A}^{(2)}$.

\vspace{5pt}
\noindent \textbf{2) Inter-Level Context Retrieval.}
To fully exploit the information provided by the coarse representation, we propose an inter-level retrieval strategy that transfers coarse-level attributes (e.g., feature $\mathbf{f}$, scaling $\mathbf{s}$, and offsets $o$) to the fine-level query anchors, establishing a preliminary feature representation for subsequent geometric context refinement. 
This strategy utilizes the voxel grid as a spatial bridge to align attributes across levels. Specifically, based on the voxel partitioning rules in Eq.~(\ref{6}), we establish a deterministic mapping $\mathcal{Q}$ that associates each fine-level query anchor $\mathbf{a}_j^{(2)} \in \mathcal{A}^{(2)}$ with its corresponding coarse-level parent $\mathbf{a}_i^{(1)}$ within the same voxel, yielding a preliminary context set $\mathcal{A}^{(2)}_{pre}$:
\begin{equation}
\mathcal{A}^{(2)}_{pre} = { \{(\mathbf{p}_j^{(2)}, \mathbf{f}_i^{(1)},\mathbf{s}_i^{(1)},\mathbf{o}_i^{(1)}) \mid \mathbf{a}_j^{(2)} \in \mathcal{A}^{(2)}, i = \mathcal{Q}(\mathbf{a}_j^{(2)}) }\},
\end{equation}
where $\mathbf{p}_j^{(2)}$ is the fine-level position and the tuple $(\mathbf{f}_i^{(1)}, \mathbf{s}_i^{(1)}, o_i^{(1)})$ denotes the decoded attributes inherited from the $i$-th coarse-level anchor in $\mathcal{A}^{(1)}$. 
To explicitly capture the geometric relations among neighboring contexts and model local structure dependencies, we construct a Ball k-nearest neighbor (Ball k-NN) graph~\cite{qi2017pointnet++} over $\mathcal{A}^{(2)}_{pre}$ based on the spatial positions of the fine-level anchors. 
This graph prior, denoted as $\mathcal{G}^{(2)}_{geo} = (\mathcal{A}^{(2)}_{pre}, \mathcal{E})$, where $\mathcal{E}$ represents the edges connecting spatially neighboring anchors, encapsulates both the inherited coarse-level preliminary contexts and the fine-level local geometric correlations. It provides a structured foundation for subsequent geometry-guided feature refinement in the GG-Conv.


\vspace{5pt}
\noindent \textbf{3) Geometry-Guided Convolution (GG-Conv).}
For the graph-based anchor prior $\mathcal{G}^{(2)}_{geo}$ with its $k$-NN graph structure, a straightforward approach is to directly process these retrieved features through a Multi-Layer Perceptron (MLP), as adopted in Context-GS \cite{wang2024ContextGS}. However, while $\mathcal{G}^{(2)}_{geo}$ provides a coarse graph-structured contextual priors, simply concatenating them cannot adequately capture the fine-grained geometric correlations in irregular and unstructured anchor distributions. To address this, we propose GG-Conv, which performs adaptive feature refinement upon the local neighborhood. As illustrated in Figure~\ref{fig:context_module}, GG-Conv refines the preliminary features of each query anchor $\mathbf{a}_{i(pre)}^{(2)}$  by aggregating context priors from its $k$-NN neighbors $\mathbf{a}_{j(pre)}^{(2)}$ through two cooperative branches. 

\textbf{Geometry Branch}: To capture local geometric correlations, we compute normalized relative offsets $\Delta \hat{\mathbf{p}}_{ij} = \text{Norm}(\mathbf{p}_j - \mathbf{p}_i)$ to query a learnable 3D kernel $\mathbf{T} \in \mathbb{R}^{D \times D \times D \times C}$. Specifically, $\Delta \hat{\mathbf{p}}_{ij}$ is treated as a continuous query coordinate within the grid space of~$\mathbf{T}$, from which a dynamic kernel $\mathbf{w}_{ij}$ is retrieved via trilinear interpolation:
\begin{equation}
\mathbf{w}_{ij} = \text{Tri-Interp}(\text{Look-up}(\mathbf{T}, \Delta \hat{\mathbf{p}}_{ij})).
\end{equation}
Here, $\text{Look-up}(\cdot)$ denotes the indexing operation that retrieves entries from the learnable geometry-aware weight table $\mathbf{T}$ based on the relative displacement $\Delta \hat{\mathbf{p}}_{ij}$, $\text{Tri-Interp}(\cdot)$ interpolates over the eight nearest integer grid cells in $\mathbf{T}$ to perceive fine-grained geometric variations beyond discrete grid resolutions. This mechanism ensures spatially continuous weighting for effective geometry-aware feature aggregation.

\textbf{Feature Branch}: Concurrently, the feature branch extracts 
semantic geometric correlations. Specifically, we first concatenate the residual features $\Delta \mathbf{f}_{ij} = \mathbf{f}_j - \mathbf{f}_i$ with the relative spatial offsets $\Delta \mathbf{p}_{ij}$. The concatenated feature is then transformed via an MLP to extract refined semantic embeddings that are implicitly aware of the local geometry:
\begin{equation}
\mathbf{e}_{ij} = MLP([\Delta \mathbf{f}_{ij}  \parallel  \Delta \mathbf{p}_{ij}])
\end{equation}where $[\cdot \parallel \cdot]$ denotes the concatenation operation.
Finally, the dynamic geometric weights from the first branch explicitly modulate these embeddings to produce the geometry-guided context $\mathbf{f}_{i(ctx)}$:
\begin{equation}
\mathbf{f}_{i(ctx)} = \sum_{j \in \mathcal{N}(i)} \mathbf{w}_{ij} \odot \mathbf{e}_{ij}
\end{equation}
where $\odot$ denotes the element-wise product. 

The Local Geometry-Guided Context $\mathbf{a}_{i(ctx)}$ is formed by concatenating the refined geometry-aware feature $\mathbf{f}_{i(ctx)}$ with inherited attributes ($\mathbf{p}_i, \mathbf{s}_i, o_i$). By integrating local neighborhood correlations with global geometric priors, $\mathbf{a}_{i(ctx)}$ provides a comprehensive representation of the anchor's state. This $\mathbf{a}_{i(ctx)}$ is subsequently projected via an MLP to predict distribution parameters:
\begin{equation}
\mu, \sigma, \Delta_{\text{adj}} = \text{MLP}(\mathbf{a}_{i(ctx)}),
\end{equation}
where $\mu$ and $\sigma$ characterize the attribute distribution, and $\Delta{\text{adj}}$ adaptively adjusts the quantization step based on local structural complexity. By operating on graph-structured priors, GG-Conv enables spatially adaptive bitrate allocation, prioritizing complex geometric regions while maintaining high compression efficiency in smoother areas.

\subsection{Bitstream Composition.}
The final bitstream is composed of  three parts: $R = R_{\text{geo}} + R_{\text{attr}} + R_{\text{model}}$, where $R_{\text{geo}}$, $R_{\text{attr}}$, and $R_{\text{model}}$ denote the bitrate associated with anchor geometry, hierarchical anchor attributes, and model parameters, respectively.
Specifically, $R_{\text{geo}}$ represents the explicit anchor coordinates, which are losslessly compressed after quantization. 
$R_{\text{attr}}$ comprises the hierarchical attributes (features, scalings, offsets) of both $\mathcal{A}^{(1)}$ and $\mathcal{A}^{(2)}$, encoded via the proposed hierarchical geometry-guided context modeling.
$R_{\text{model}}$ stores the quantized weights of the shared MLPs and GG-Conv modules as a lightweight model header.

\subsection{Optimization}
The proposed LG-HCC for 3DGS compression is optimized under a rate-distortion objective:
\begin{equation}
    \mathcal{L} = \mathcal{L}_{\text{render}} + \lambda \, \mathcal{L}_{\text{anchor}},
    \label{eq:total_loss}
\end{equation}
where
$ \mathcal{L}_{\text{render}} $  denotes the rendering loss inherited from Scaffold-GS~\cite{lu2024scaffold} and serves as the distortion term, while $ \mathcal{L}_{\text{anchor}} $
denotes the estimated entropy-coded bitrate of the anchor attributes, including positions, opacities, and feature descriptors, following HAC++~\cite{chen2025hac++} and Context-GS~\cite{wang2024ContextGS}, and serves as the rate term.
The hyperparameter  $ \lambda > 0 $  controls the trade-off between reconstruction fidelity ( $ \mathcal{L}_{\text{render}} $ ) and compression efficiency ( $ \mathcal{L}_{\text{anchor}} $ ).

\begin{figure*}[!t]
\centering
\includegraphics[width=1\textwidth]{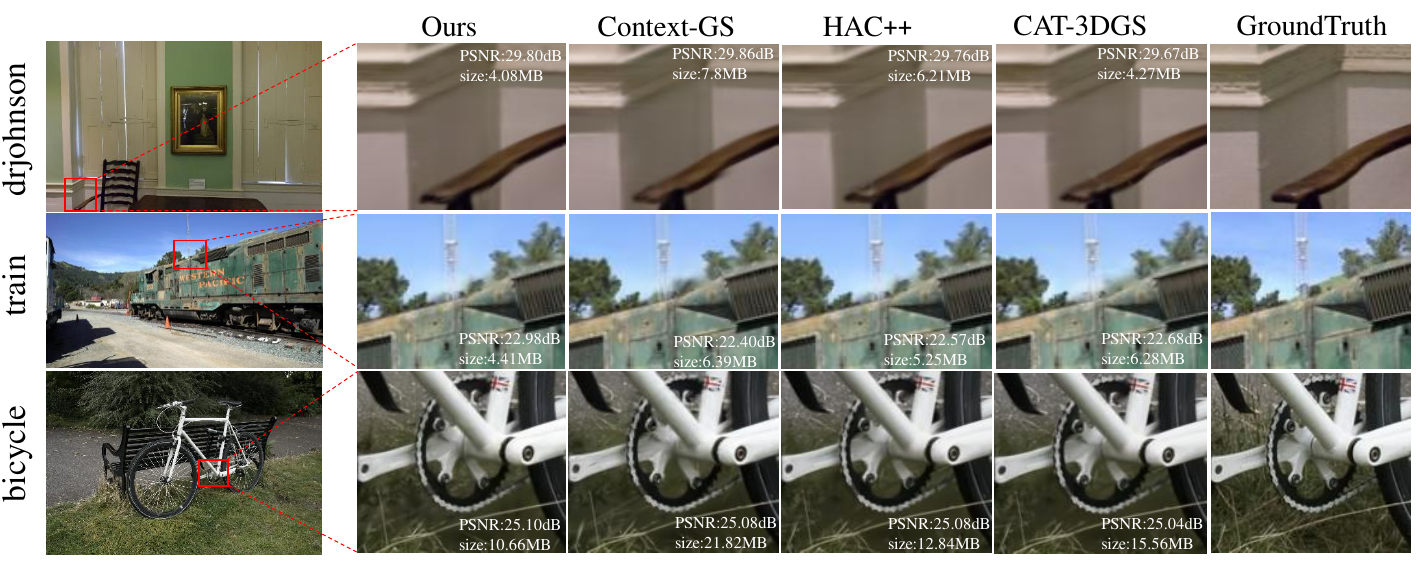} 
\caption{Qualitative results of the proposed LG-HCC method compared to existing three compression methods on the drjohnson (DeepBlending), train (Tank\&Temples) and bicycile (Mip-NeRF360) scenes, respectively.}
\label{fig:visual_comparison}
\end{figure*}

\definecolor{bestcell}{HTML}{FFC7CE}   
\definecolor{secondcell}{HTML}{FFF2CC} 
\begin{table*}[t]
\centering
\caption{\textbf{Comparison against existing compression approaches.}
Our method is evaluated under two settings: a \textbf{high-fidelity mode} and a \textbf{high-compression mode}, showcasing its versatility.
Cells colored \colorbox{bestcell}{Red} and \colorbox{secondcell}{yellow} highlight the best-performing and second-best results, respectively, for both the low-bitrate and high-bitrate regimes.
All size are reported in MB.}
\label{tab:quantitative}

\setlength{\tabcolsep}{0.5pt}

\begin{tabular}{l|cccc|cccc|cccc|cccc}
\toprule
\multirow{1}{*}{\textbf{Datasets}} & \multicolumn{4}{c|}{\textbf{Mip-NeRF360}~\cite{barron2022mip}} & \multicolumn{4}{c|}{\textbf{BungeeNeRF}~\cite{xiangli2022bungeenerf}} & \multicolumn{4}{c|}{\textbf{DeepBlending}~\cite{hedman2018deep}} & \multicolumn{4}{c}{\textbf{Tank\&Temples}~\cite{knapitsch2017tanks}} \\
 \multirow{1}{*}{\textbf{Methods}}& \scriptsize{PSNR(dB)}$\uparrow$ & \footnotesize{SSIM}$\uparrow$ & \footnotesize{LPIPS}$\downarrow$ & size$\downarrow$ & \scriptsize{PSNR(dB)}$\uparrow$ & \footnotesize{SSIM}$\uparrow$ & \footnotesize{LPIPS}$\downarrow$ & size$\downarrow$ & \scriptsize{PSNR(dB)}$\uparrow$ & \footnotesize{SSIM}$\uparrow$ & \footnotesize{LPIPS}$\downarrow$ & size$\downarrow$ & \scriptsize{PSNR(dB)}$\uparrow$ & \footnotesize{SSIM}$\uparrow$ & \footnotesize{LPIPS}$\downarrow$ & size$\downarrow$ \\
\midrule
3DGS~\cite{bagdasarian24073dgs} & 27.49 & \colorbox{bestcell}{0.813} &  c{0.222} & 744.7 & 24.87 & 0.841 & 0.205 & 1616 & 29.42 & 0.899 &  \colorbox{bestcell}{0.247} & 663.9 & 23.69 & 0.844 & 0.178 & 431.0 \\
Scaffold-GS~\cite{lu2024scaffold} & 27.50 & 0.806 & 0.252 & 253.9 & 26.62 & 0.865 & 0.241 & 183.0 & 30.21 & 0.906 & 0.254 & 66.00 & 23.96 & 0.853 & 0.177 & 86.50 \\
\midrule
Compact3DGS~\cite{lee2024compact} & 27.08 & 0.798 & 0.247 & 48.80 & 23.36 & 0.788 & 0.251 & 82.60 & 29.79 & 0.901 & 0.258 & 43.21 & 23.32 & 0.831 & 0.201 & 39.43 \\
Compressed3D~\cite{navaneet2023compact3d} & 26.98 & 0.801 & 0.238 & 28.80 & 24.13 & 0.802 & 0.245 & 55.79 & 29.38 & 0.898 & 0.253 & 25.30 & 23.32 & 0.832 & 0.194 & 17.28 \\
Morgen. et al.~\cite{morgenstern2024compact} & 26.01 & 0.772 & 0.259 & 23.90 & 22.43 & 0.708 & 0.339 & 48.25 & 28.92 & 0.891 & 0.276 & 8.40 & 22.78 & 0.817 & 0.211 & 13.05 \\
CompGS~\cite{liu2024compgs} & 27.26 & 0.803 & 0.239 &16.50 & - & - & - & - & 29.69 & 0.901 & 0.279 & 8.77 & 23.70 & 0.837 & 0.208 & 9.60 \\
\midrule

HAC (low-rate)~\cite{chen2024hac} &27.53 & {0.807} & {0.238} & 15.26 & 26.48 & 0.845 & 0.25 & 18.49 & 29.98 & 0.902 & 0.269 & 4.35 & 24.04 & 0.846 & 0.187 & 8.10 \\
HAC (high-rate)~\cite{chen2024hac} & {27.77} &{0.811} &  \colorbox{secondcell}{0.230} & 21.84 & 27.08 & 0.872 & 0.209 & 29.72 & 30.34 & 0.906 & 0.258 & 6.35 &  24.40 & 0.853 & 0.177 & 11.24 \\
ContextGS (low-rate)~\cite{wang2024ContextGS} &  {27.62} & 0.778 & {0.237} & 12.68 & {26.90} & {0.866} & {0.222} &  14.00 &  30.11 & {0.907} & {0.265} & 3.43 & 24.20 & {0.852} & {0.184} & 7.05 \\
ContextGS (high-rate)~\cite{wang2024ContextGS} & 27.75 & {0.811} & {0.231} & 18.41 &  27.15 &  0.875 &  0.205 &  21.80 &  30.39 & 0.909 & 0.258 & 6.60 & 24.29 & 0.855 & \colorbox{secondcell}{0.176} & 11.80 \\
HAC++ (low-rate)~\cite{chen2025hac++} & 27.6 & 0.803 &  0.253 & \colorbox{secondcell}{8.34} & 26.78 & {0.858} & 0.235 & \colorbox{secondcell}{11.75} & {30.16} & {0.907} & {0.266} & \colorbox{secondcell}{2.91} & {24.22} & {0.849} & 0.190 & \colorbox{secondcell}{5.18} \\
HAC++ (high-rate)~\cite{chen2025hac++} &  \colorbox{secondcell}{27.82} & {0.811} &  {0.231} & 18.48 & 27.17 & 0.879 & \colorbox{secondcell}{0.196} & {20.82} & 30.34 & \colorbox{secondcell}{0.911} & {0.254} & 5.287 &  24.32 & 0.854 & 0.178 & {8.63} \\
CAT-3DGS~\cite{zhan2025cat} & {27.77} &  0.809 & 0.241 & {12.35} &  \colorbox{secondcell}{27.35} &  \colorbox{bestcell}{0.886} &  \colorbox{bestcell}{0.183} &  26.59 &  30.29 & 0.909 & 0.269 & {3.56} & \colorbox{secondcell}{24.41} & 0.853 & 0.189 & 6.93 \\
MoP~\cite{liu20253d} & 27.68 &  0.808 & 0.234 & 15.64 &  27.26 &  0.875 &  0.207 &  20.83 &\colorbox{secondcell}{30.45} & 0.912 & 0.250 & 5.65 & 24.21 & \colorbox{bestcell}{0.861} & \colorbox{bestcell}{0.163} & 8.98 \\
\midrule[1pt]

\textbf{Ours} (low-rate) & {27.64} & {0.805} & 0.247 &  \colorbox{bestcell}{8.23} & {26.93} &  {0.866} & {0.231} &  \colorbox{bestcell}{11.59} & {30.25} & {0.909} & 0.267 &  \colorbox{bestcell}{2.83} & {24.32} &  {0.852} & {0.187} &  \colorbox{bestcell}{4.92} \\

\textbf{Ours} (high-rate) &  \colorbox{bestcell}{27.82} & \colorbox{secondcell}{0.812} & 0.2351 & {12.24} & \colorbox{bestcell}{27.46} &  \colorbox{secondcell}{0.883} & \colorbox{secondcell}{0.196} &  {19.48} & \colorbox{bestcell}{30.49} & \colorbox{bestcell}{0.912} & \colorbox{secondcell}{0.250} & {5.63} & \colorbox{bestcell}{24.43} &  \colorbox{secondcell}{0.856} & 0.177 &  {8.49} \\

\bottomrule[1.2pt]
\end{tabular}%
\end{table*}

\begin{figure*}[!t]
\centering
\includegraphics[width=0.9\textwidth]{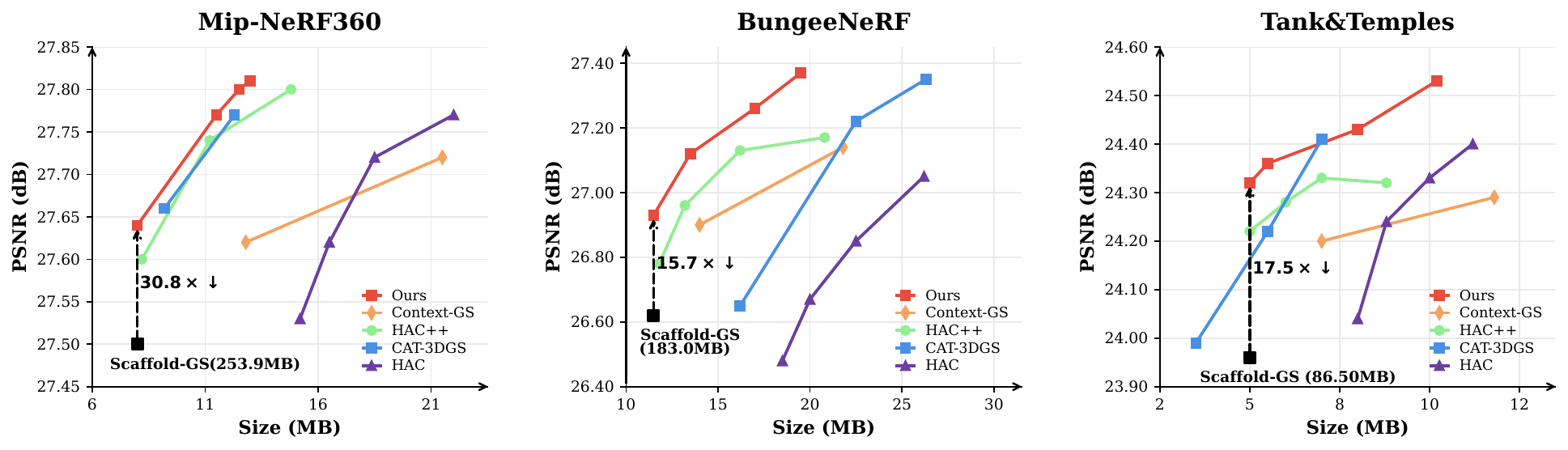} 
\caption{\textbf{Rate-Distortion (RD) performance curves.}
Comparison with advanced compression techniques (HAC, HAC++, Context-GS, CAT-3DGS) on the Mip-NeRF360, BungeeNeRF, and Tank\&Temples datasets.
LG-HCC demonstrates robust coding efficiency, maintaining the upper envelope across a wide range of bitrates.}
\label{fig:rd_curves}
\end{figure*}

\section{Experiments}
\subsection{Implementation Details}
We implement LG-HCC within the PyTorch framework, building upon the official Scaffold-GS codebase~\cite{lu2024scaffold}. To ensure the convergence of geometry-aware components, models are trained for 35,000 iterations. We adopt a minimalist two-level hierarchy ($L=2$) and set the voxelization scaling factor $s$ consistent with Context-GS~\cite{wang2024ContextGS}. For local geometry-aware perception via Ball K-NN graph construction, the neighbor count is uniformly fixed at $K=8$ for both NAAP and the formation of $\mathcal{G}^{(2)}_{geo}$, while the GG-Conv volumetric kernel size is set to $k=5$. Unless otherwise specified, we follow the default Scaffold-GS configuration, utilizing an anchor feature dimension of 50 and a compressed latent dimension of 12.


\begin{table*}[!t]
\centering
\caption{Comparison of storage breakdown (Positions, Features, MLP, others; "others" includes auxiliary structures and metadata (e.g., scalings, offsets, masks).),  and rendering FPS on the "Bicycle" scene between Context-GS (vanilla EM, i.e., Entropy Modeling) and our GG-Conv-based Hierarchical Geometry-Guided Context EM.}
\label{tab:GG-Conv}
\small
\setlength{\tabcolsep}{5pt}
\begin{tabular}{lcccccccccc}
\toprule
\multirow{2}{*}{Methods} & \multicolumn{5}{c}{Storage Costs (MB)$\downarrow$} & \multicolumn{3}{c}{Fidelity} & \multicolumn{1}{c}{Rendering} \\

\cmidrule(lr){2-6} \cmidrule(lr){7-9} \cmidrule(lr){10-10}

& Positions & Features & MLPs & Others & Total & PSNR$\uparrow$ & SSIM$\uparrow$ & LPIPS$\downarrow$ &   FPS \\
\midrule

Vanilla EM~\cite{wang2024ContextGS} &5.24 & 8.27 & 0.3162 & 11.24 & 25.07 &25.07 & 0.7391 & 0.2683 &135 \\
\textbf{GG-Conv-based EM (two level)} & \textbf{4.69} & \textbf{4.51} & \textbf{0.2381}  &\textbf{6.31} & \textbf{15.75}& \textbf{25.16} & \textbf{0.7415} & \textbf{0.2704} &\textbf{170} \\

GG-Conv-based EM (three level) & 4.66 & 6.82 & 0.3162 & 7.54 & 19.34 & 25.10 & 0.7416 & 0.2703 &160 \\
\bottomrule
\end{tabular}
\end{table*}

\subsection{Experiment Details}
\vspace{5pt}
\textbf{Datasets.}
Following previous methods~\cite{chen2025hac++,zhan2025cat,wang2024ContextGS}, our method is evaluated on four standard real-world benchmarks: Mip-NeRF360~\cite{barron2022mip} (utilizing all 9 scenes), BungeeNeRF~\cite{xiangli2022bungeenerf}, DeepBlending~\cite{hedman2018deep}, and Tanks\&Temples~\cite{knapitsch2017tanks}.
\vspace{5pt}

\noindent \textbf{Baseline Methods.}
We compare our method with a wide range of state-of-the-art 3DGS compression approaches, which can be grouped into two main streams. The first focuses on compact representations via parameter pruning or vector quantization, including Scaffold-GS~\cite{lu2024scaffold}, Compressed3D~\cite{navaneet2023compact3d}, and CompGS~\cite{liu2024compgs}.
The second stream, more related to our work, improves entropy coding by exploiting contextual priors. Examples include HAC~\cite{chen2024hac} and HAC++~\cite{chen2025hac++}, which adopt hash grids for spatial compactness, and Context-GS~\cite{wang2024ContextGS}, which models anchor-level context as hyperpriors. More recently, MoP~\cite{liu20253d} introduced a Mixture-of-Experts (MoE) module for robust feature learning, while CAT-3DGS~\cite{zhan2025cat3dgs} applies channel-wise autoregressive modeling for attribute compression.
We benchmark against these methods to validate the effectiveness of our local geometry-aware hierarchical context modeling.

\noindent \textbf{Metrics.}
We evaluate compression performance in terms of storage
size, measured in megabytes (MB). To assess the visual quality of rendered images generated from the compressed 3DGS data, we
employ three standard metrics: Peak Signal-to-Noise Ratio (PSNR),
Structural Similarity Index (SSIM)~\cite{1284395}, and Learned Perceptual Image
Patch Similarity (LPIPS)~\cite{zhang2018unreasonable}.


\subsection{Experiment Results}
\vspace{5pt}
\textbf{Quantitative Analysis.}
As shown in Table~\ref{tab:quantitative}, our proposed LG-HCC achieves significant improvements over the Scaffold-GS~\cite{lu2024scaffold} baseline. Specifically, it achieves up to a \textbf{30.8$\times$} storage reduction on the Mip-NeRF360 dataset, while maintaining competitive or even superior rendering fidelity in terms of PSNR and SSIM. Compared with state-of-the-art methods, including HAC~\cite{chen2024hac}, HAC++~\cite{chen2025hac++}, Context-GS~\cite{wang2024ContextGS}, CAT-3DGS~\cite{zhan2025cat}, and MoP~\cite{liu20253d}, our LG-HCC consistently achieves better rate-distortion performance, especially at low-rate.
Notably, in several scenes under high-rate settings, LG-HCC even surpasses the uncompressed Scaffold-GS in both PSNR and SSIM. 
Furthermore, we illustrate the Rate-Distortion curves in Figure~\ref{fig:rd_curves} to confirm that LG-HCC provides a superior performance envelope, achieving higher rendering quality at equivalent bitrates across a wide range of compression ratios on four representative datasets.

\vspace{5pt}
\noindent \textbf{Qualitative Analysis.}
As shown in Figure~\ref{fig:visual_comparison}, LG-HCC produces sharper structural details and significantly fewer artifacts compared to representative anchor-based methods. This improvement stems from our locally geometry-aware constraints, which leverage local geometric context as strong regularization to suppress floaters while enhancing fine local texture details. Thanks to the geometry-guided context modeling via GG-Conv, our method effectively preserves high-frequency textures even at low bitrates, where other compression approaches often suffer from blurring or ``popping'' artifacts.

\begin{table}[h]
\centering
\caption{Quantitative ablation study on the Deep Blending dataset. 
  }
\label{tab:ablation-deep-blending}
\small
\setlength{\tabcolsep}{4pt}
\begin{tabular}{lcccc}
\toprule
Method & \footnotesize{PSNR (dB)}$\uparrow$ & \footnotesize{SSIM}$\uparrow$ & \footnotesize{LPIPS}$\downarrow$ & size (\footnotesize{MB})$\downarrow$ \\
\midrule
Ours w/o NAAP \& GG-Conv   
    & 30.10 & 0.9062 & 0.2659 & 3.76 \\
Ours w/o GG-Conv                  
    & 30.13 & 0.9087 & 0.2656 & 3.72 \\
Ours w/o NAAP                      
    & 30.20 & 0.9095 & 0.2648 & 3.58 \\
\textbf{Ours (full)}  
    & \textbf{30.31} & \textbf{0.9106} & \textbf{0.2636} & \textbf{3.47} \\
\bottomrule
\end{tabular}
\end{table}

\subsection{Ablation Study and Analysis}
\vspace{5pt}
\noindent \textbf{Effectiveness of Designed Components.}
To evaluate the efficacy of each component in our LG-HCC framework, we conduct an ablation study on the Deep Blending dataset considering four configurations: (1) Ours (full): the full model with both NAAP and GG-Conv; (2) Ours w/o GG-Conv: the model excluding the entropy modeling component; (3) Ours w/o NAAP: the model without the graph-based pruning strategy; and (4) Ours w/o NAAP \& GG-Conv: the baseline version without both proposed modules.

As reported in Table~\ref{tab:ablation-deep-blending}, the full model consistently outperforms all other variants. Specifically, removing NAAP ("Ours w/o NAAP") incurs a PSNR drop of 0.11 dB and a slight storage increase of 0.11 MB, validating its effectiveness in redundancy removal. The contribution of GG-Conv is even more pronounced: its absence ("Ours w/o GG-Conv") results in a significant performance degradation of 0.18 dB in PSNR and an additional 0.25 MB in bit-rate. When both modules are deactivated ("Ours w/o NAAP \& GG-Conv"), the rendering quality drops by 0.21 dB, emphasizing their collective importance for high-fidelity reconstruction under extreme compression.
Notably, the combined performance gain from NAAP and GG-Conv exceeds the sum of their individual improvements, revealing strong synergy between the two modules.
This indicates that NAAP’s preservation of local geometric structures among anchors creates more structured representations that enhance the effectiveness of subsequent hierarchical entropy coding, allowing GG-Conv to achieve superior compression efficiency and rendering quality under tight bit constraints.

\vspace{5pt}
\noindent \textbf{Effectiveness of NAAP.} As shown in Table~\ref{tab:naap-vs-pruning} on the Deep Blending dataset, compared to vanilla pruning~\cite{wang2024ContextGS}, NAAP increases positions storage by only 2.1\% while reducing features storage by 2.4\%.
This indicates that NAAP selectively retains a small number of critical anchors to preserve local geometric structures. These anchors enable more coherent local geometry in the gaussian field, which in turn creates stronger spatial correlations among features. End-to-end optimization then exploits this structure to remove redundancy more effectively, allowing significantly better entropy coding of the features.
Thus, the tiny anchor overhead is largely compensated by much greater feature compression, revealing that geometry-aware anchor preservation is a highly efficient way to improve structured representation and overall compression in Gaussian splatting.

\vspace{5pt}
\noindent \textbf{Effectiveness of Hierarchical Geometry-Guided Context Modeling.}
We evaluate the proposed hierarchical geometry-guided context modeling module against the vanilla entropy model~\cite{wang2024ContextGS} under the same NAAP framework. As shown in Table~\ref{tab:GG-Conv}, compared to the baseline, our two-level GG-Conv-based EM significantly reduces the total storage cost from 25.07 MB to 15.75 MB (37.2\% reduction) while improving PSNR from 25.07 dB to 25.16 dB. Most notably, the storage for anchor features is nearly halved (from 8.27 MB to 4.51 MB), and the overhead in the "Others" category drops from 11.24 MB to 6.31 MB, demonstrating superior redundancy elimination.
We also find that deeper hierarchies do not necessarily enhance efficiency. Although the three-level scheme slightly optimizes position storage (4.66 MB), it increases feature costs to 6.82 MB and total storage to 19.34 MB. This stems from the excessive computational overhead and complex cross-layer dependencies that provide diminishing returns in fidelity. Therefore, we adopt the two-level design.
The efficiency of our model is driven by the GG-Conv operator, which leverages local geometric correlations to dynamically weight anchor attributes, producing geometry-aware context. This facilitates the derivation of informed conditional priors with high structural fidelity, which enhances the exploitation of inter-anchor dependencies and ensures a thorough reduction of attribute redundancy in both features and metadata.

\begin{table}[h]
\centering
\caption{Storage comparison between vanilla pruning (Context-GS~\cite{wang2024ContextGS}) and our LG-HCC method.}
\label{tab:naap-vs-pruning}
\small
\setlength{\tabcolsep}{5pt}  
\begin{tabular}{lcc}
\toprule
Method & Positions (MB)$\downarrow$ & Features (MB)$\downarrow$ \\
\midrule
Vanilla Pruning & 0.7531 & 1.1108 \\
\textbf{NAAP (Ours)}     & \textbf{0.7689} \,(+2.1\%) & \textbf{1.0847} \,(-2.4\%) \\
\bottomrule
\end{tabular}
\end{table}

\section{Conclusion}
To enhance 3DGS compression performance, we propose a local geometry-aware Hierarchical Context Compression framework that consistently exploits geometric relationships within anchor neighborhoods. By introducing Neighborhood-Aware Anchor Pruning (NAAP), we achieve a sparse yet locally geometry-consistent hierarchy through adaptive neighbor merging. Moreover, a hierarchical geometry-guided context modeling scheme is designed that is powered by the lightweight GG-Conv operator and leverages geometry-guided coarse-to-fine context priors to effectively minimize inter-anchor redundancy. Extensive experiments demonstrate that our method achieves state-of-the-art rate-distortion performance, delivering superior rendering quality at substantially reduced storage costs.


\bibliographystyle{ACM-Reference-Format}
\bibliography{sample-base}

@String{Computer = "{IEEE} Computer" }

@String{Springer = "Springer-Verlag" }

@article{kerbl20233d,
  title={3D Gaussian splatting for real-time radiance field rendering.},
  author={Kerbl, Bernhard and Kopanas, Georgios and Leimk{\"u}hler, Thomas and Drettakis, George},
  journal={ACM Trans. Graph.},
  volume={42},
  number={4},
  pages={139--1},
  year={2023}
}

@inproceedings{lee2024compact,
  title={Compact 3d gaussian representation for radiance field},
  author={Lee, Joo Chan and Rho, Daniel and Sun, Xiangyu and Ko, Jong Hwan and Park, Eunbyung},
  booktitle={Proceedings of the IEEE/CVF Conference on Computer Vision and Pattern Recognition},
  pages={21719--21728},
  year={2024}
}

@inproceedings{niedermayr2024compressed,
  title={Compressed 3d gaussian splatting for accelerated novel view synthesis},
  author={Niedermayr, Simon and Stumpfegger, Josef and Westermann, R{\"u}diger},
  booktitle={Proceedings of the IEEE/CVF Conference on Computer Vision and Pattern Recognition},
  pages={10349--10358},
  year={2024}
}

@inproceedings{girish2024eagles,
  title={Eagles: Efficient accelerated 3d gaussians with lightweight encodings},
  author={Girish, Sharath and Gupta, Kamal and Shrivastava, Abhinav},
  booktitle={European Conference on Computer Vision},
  pages={54--71},
  year={2024},
  organization={Springer}
}

@article{fan2024lightgaussian,
  title={Lightgaussian: Unbounded 3d gaussian compression with 15x reduction and 200+ fps},
  author={Fan, Zhiwen and Wang, Kevin and Wen, Kairun and Zhu, Zehao and Xu, Dejia and Wang, Zhangyang and others},
  journal={Advances in neural information processing systems},
  volume={37},
  pages={140138--140158},
  year={2024}
}

@inproceedings{cheng2020learned,
  title={Learned image compression with discretized gaussian mixture likelihoods and attention modules},
  author={Cheng, Zhengxue and Sun, Heming and Takeuchi, Masaru and Katto, Jiro},
  booktitle={Proceedings of the IEEE/CVF conference on computer vision and pattern recognition},
  pages={7939--7948},
  year={2020}
}

@inproceedings{he2021checkerboard,
  title={Checkerboard context model for efficient learned image compression},
  author={He, Dailan and Zheng, Yaoyan and Sun, Baocheng and Wang, Yan and Qin, Hongwei},
  booktitle={Proceedings of the IEEE/CVF Conference on Computer Vision and Pattern Recognition},
  pages={14771--14780},
  year={2021}
}

@inproceedings{lu2024scaffold,
  title={Scaffold-gs: Structured 3d gaussians for view-adaptive rendering},
  author={Lu, Tao and Yu, Mulin and Xu, Linning and Xiangli, Yuanbo and Wang, Limin and Lin, Dahua and Dai, Bo},
  booktitle={Proceedings of the IEEE/CVF Conference on Computer Vision and Pattern Recognition},
  pages={20654--20664},
  year={2024}
}

@article{mildenhall2021nerf,
  title={Nerf: Representing scenes as neural radiance fields for view synthesis},
  author={Mildenhall, Ben and Srinivasan, Pratul P and Tancik, Matthew and Barron, Jonathan T and Ramamoorthi, Ravi and Ng, Ren},
  journal={Communications of the ACM},
  volume={65},
  number={1},
  pages={99--106},
  year={2021},
  publisher={ACM New York, NY, USA}
}

@article{muller2022instant,
  title={Instant neural graphics primitives with a multiresolution hash encoding},
  author={M{\"u}ller, Thomas and Evans, Alex and Schied, Christoph and Keller, Alexander},
  journal={ACM transactions on graphics (TOG)},
  volume={41},
  number={4},
  pages={1--15},
  year={2022},
  publisher={ACM New York, NY, USA}
}

@inproceedings{chen2024hac,
  title={Hac: Hash-grid assisted context for 3d gaussian splatting compression},
  author={Chen, Yihang and Wu, Qianyi and Lin, Weiyao and Harandi, Mehrtash and Cai, Jianfei},
  booktitle={European Conference on Computer Vision},
  pages={422--438},
  year={2024},
  organization={Springer}
}

@article{chen2025hac++,
  title={Hac++: Towards 100x compression of 3d gaussian splatting},
  author={Chen, Yihang and Wu, Qianyi and Lin, Weiyao and Harandi, Mehrtash and Cai, Jianfei},
  journal={arXiv preprint arXiv:2501.12255},
  year={2025}
}

@article{wang2024ContextGS,
  title={Contextgs: Compact 3d gaussian splatting with anchor level context model},
  author={Wang, Yufei and Li, Zhihao and Guo, Lanqing and Yang, Wenhan and Kot, Alex and Wen, Bihan},
  journal={Advances in neural information processing systems},
  volume={37},
  pages={51532--51551},
  year={2024}
}

@article{zhan2025cat,
  title={CAT-3DGS: A context-adaptive triplane approach to rate-distortion-optimized 3DGS compression},
  author={Zhan, Yu-Ting and Ho, Cheng-Yuan and Yang, Hebi and Chen, Yi-Hsin and Chiang, Jui Chiu and Liu, Yu-Lun and Peng, Wen-Hsiao},
  journal={arXiv preprint arXiv:2503.00357},
  year={2025}
}

@inproceedings{liu20253d,
  title={3D Gaussian Splatting Data Compression with Mixture of Priors},
  author={Liu, Lei and Chen, Zhenghao and Xu, Dong},
  booktitle={Proceedings of the 33rd ACM International Conference on Multimedia},
  pages={8341--8350},
  year={2025}
}

@article{bagdasarian24073dgs,
  title={3dgs. zip: a survey on 3D gaussian splatting compression methods (2024)},
  author={Bagdasarian, Milena T and Knoll, Paul and Barthel, Florian and Hilsmann, Anna and Eisert, Peter and Morgenstern, Wieland},
  journal={URL https://arxiv. org/abs/2407.09510}
}

@article{wu2024recent,
  title={Recent advances in 3d gaussian splatting},
  author={Wu, Tong and Yuan, Yu-Jie and Zhang, Ling-Xiao and Yang, Jie and Cao, Yan-Pei and Yan, Ling-Qi and Gao, Lin},
  journal={Computational Visual Media},
  volume={10},
  number={4},
  pages={613--642},
  year={2024},
  publisher={TUP}
}

@article{navaneet2023compact3d,
  title={Compact3d: Compressing gaussian splat radiance field models with vector quantization},
  author={Navaneet, K and Meibodi, Kossar Pourahmadi and Koohpayegani, Soroush Abbasi and Pirsiavash, Hamed},
  journal={arXiv preprint arXiv:2311.18159},
  volume={2},
  number={3},
  year={2023}
}

@article{papantonakis2024reducing,
  title={Reducing the memory footprint of 3d gaussian splatting},
  author={Papantonakis, Panagiotis and Kopanas, Georgios and Kerbl, Bernhard and Lanvin, Alexandre and Drettakis, George},
  journal={Proceedings of the ACM on Computer Graphics and Interactive Techniques},
  volume={7},
  number={1},
  pages={1--17},
  year={2024},
  publisher={ACM New York, NY, USA}
}

@article{ali2024trimming,
  title={Trimming the fat: Efficient compression of 3d gaussian splats through pruning},
  author={Ali, Muhammad Salman and Qamar, Maryam and Bae, Sung-Ho and Tartaglione, Enzo},
  journal={arXiv preprint arXiv:2406.18214},
  year={2024}
}

@article{ortega2018graph,
  title={Graph signal processing: Overview, challenges, and applications},
  author={Ortega, Antonio and Frossard, Pascal and Kova{\v{c}}evi{\'c}, Jelena and Moura, Jos{\'e} MF and Vandergheynst, Pierre},
  journal={Proceedings of the IEEE},
  volume={106},
  number={5},
  pages={808--828},
  year={2018},
  publisher={IEEE}
}

@inproceedings{wang2024end,
  title={End-to-end rate-distortion optimized 3d gaussian representation},
  author={Wang, Henan and Zhu, Hanxin and He, Tianyu and Feng, Runsen and Deng, Jiajun and Bian, Jiang and Chen, Zhibo},
  booktitle={European Conference on Computer Vision},
  pages={76--92},
  year={2024},
  organization={Springer}
}

@inproceedings{morgenstern2024compact,
  title={Compact 3d scene representation via self-organizing gaussian grids},
  author={Morgenstern, Wieland and Barthel, Florian and Hilsmann, Anna and Eisert, Peter},
  booktitle={European Conference on Computer Vision},
  pages={18--34},
  year={2024},
  organization={Springer}
}

@article{wang2019dynamic,
  title={Dynamic graph cnn for learning on point clouds},
  author={Wang, Yue and Sun, Yongbin and Liu, Ziwei and Sarma, Sanjay E and Bronstein, Michael M and Solomon, Justin M},
  journal={ACM Transactions on Graphics (tog)},
  volume={38},
  number={5},
  pages={1--12},
  year={2019},
  publisher={Acm New York, NY, USA}
}

@inproceedings{qi2017pointnet,
  title={Pointnet: Deep learning on point sets for 3d classification and segmentation},
  author={Qi, Charles R and Su, Hao and Mo, Kaichun and Guibas, Leonidas J},
  booktitle={Proceedings of the IEEE conference on computer vision and pattern recognition},
  pages={652--660},
  year={2017}
}

@article{qi2017pointnet++,
  title={Pointnet++: Deep hierarchical feature learning on point sets in a metric space},
  author={Qi, Charles Ruizhongtai and Yi, Li and Su, Hao and Guibas, Leonidas J},
  journal={Advances in neural information processing systems},
  volume={30},
  year={2017}
}

@article{deng2025pvinet,
  title={PVINet: Point-Voxel Interlaced Network for Point Cloud Compression},
  author={Deng, Xuan and Wang, Xingtao and Meng, Xiandong and Zhao, Debin and Fan, Xiaopeng},
  journal={IEEE Signal Processing Letters},
  volume={33},
  pages={61--65},
  year={2025},
  publisher={IEEE}
}

@article{fan2022d,
  title={D-dpcc: Deep dynamic point cloud compression via 3d motion prediction},
  author={Fan, Tingyu and Gao, Linyao and Xu, Yiling and Li, Zhu and Wang, Dong},
  journal={arXiv preprint arXiv:2205.01135},
  year={2022}
}

@article{yang2024spectrally,
  title={Spectrally pruned gaussian fields with neural compensation},
  author={Yang, Runyi and Zhu, Zhenxin and Jiang, Zhou and Ye, Baijun and Chen, Xiaoxue and Zhang, Yifei and Chen, Yuantao and Zhao, Jian and Zhao, Hao},
  journal={arXiv preprint arXiv:2405.00676},
  year={2024}
}

@inproceedings{liu2024compgs,
  title={Compgs: Efficient 3d scene representation via compressed gaussian splatting},
  author={Liu, Xiangrui and Wu, Xinju and Zhang, Pingping and Wang, Shiqi and Li, Zhu and Kwong, Sam},
  booktitle={Proceedings of the 32nd ACM International Conference on Multimedia},
  pages={2936--2944},
  year={2024}
}

@inproceedings{barron2022mip,
  title={Mip-nerf 360: Unbounded anti-aliased neural radiance fields},
  author={Barron, Jonathan T and Mildenhall, Ben and Verbin, Dor and Srinivasan, Pratul P and Hedman, Peter},
  booktitle={Proceedings of the IEEE/CVF conference on computer vision and pattern recognition},
  pages={5470--5479},
  year={2022}
}

@inproceedings{xiangli2022bungeenerf,
  title={Bungeenerf: Progressive neural radiance field for extreme multi-scale scene rendering},
  author={Xiangli, Yuanbo and Xu, Linning and Pan, Xingang and Zhao, Nanxuan and Rao, Anyi and Theobalt, Christian and Dai, Bo and Lin, Dahua},
  booktitle={European conference on computer vision},
  pages={106--122},
  year={2022},
  organization={Springer}
}

@article{hedman2018deep,
  title={Deep blending for free-viewpoint image-based rendering},
  author={Hedman, Peter and Philip, Julien and Price, True and Frahm, Jan-Michael and Drettakis, George and Brostow, Gabriel},
  journal={ACM Transactions on Graphics (ToG)},
  volume={37},
  number={6},
  pages={1--15},
  year={2018},
  publisher={ACM New York, NY, USA}
}

@article{knapitsch2017tanks,
  title={Tanks and temples: Benchmarking large-scale scene reconstruction},
  author={Knapitsch, Arno and Park, Jaesik and Zhou, Qian-Yi and Koltun, Vladlen},
  journal={ACM Transactions on Graphics (ToG)},
  volume={36},
  number={4},
  pages={1--13},
  year={2017},
  publisher={ACM New York, NY, USA}
}

@inproceedings{zhan2025cat3dgs,
  author    = {Yu-Ting Zhan and Cheng-Yuan Ho and Hebi Yang and Yi-Hsin Chen and Jui Chiu Chiang and Yu-Lun Liu and Wen-Hsiao Peng},
  title     = {{CAT-3DGS: A context-adaptive triplane approach to rate-distortion-optimized 3DGS compression}},
  booktitle = {Proceedings of the Thirteenth International Conference on Learning Representations (ICLR)},
  year      = {2025},
}

@ARTICLE{1284395,
  author={Zhou Wang and Bovik, A.C. and Sheikh, H.R. and Simoncelli, E.P.},
  journal={IEEE Transactions on Image Processing}, 
  title={Image quality assessment: from error visibility to structural similarity}, 
  year={2004},
  volume={13},
  number={4},
  pages={600-612},
  doi={10.1109/TIP.2003.819861}}

@inproceedings{zhang2018unreasonable,
  title={The unreasonable effectiveness of deep features as a perceptual metric},
  author={Zhang, Richard and Isola, Phillip and Efros, Alexei A and Shechtman, Eli and Wang, Oliver},
  booktitle={Proceedings of the IEEE conference on computer vision and pattern recognition},
  pages={586--595},
  year={2018}
}

@article{shin2025locality,
  title={Locality-aware gaussian compression for fast and high-quality rendering},
  author={Shin, Seungjoo and Park, Jaesik and Cho, Sunghyun},
  journal={arXiv preprint arXiv:2501.05757},
  year={2025}
}

@inproceedings{fang2024mini,
  title={Mini-splatting: Representing scenes with a constrained number of gaussians},
  author={Fang, Guangchi and Wang, Bing},
  booktitle={European conference on computer vision},
  pages={165--181},
  year={2024},
  organization={Springer}
}

@inproceedings{mallick2024taming,
  title={Taming 3dgs: High-quality radiance fields with limited resources},
  author={Mallick, Saswat Subhajyoti and Goel, Rahul and Kerbl, Bernhard and Steinberger, Markus and Carrasco, Francisco Vicente and De La Torre, Fernando},
  booktitle={SIGGRAPH Asia 2024 Conference Papers},
  pages={1--11},
  year={2024}
}

@inproceedings{kim2024color,
  title={Color-cued efficient densification method for 3d gaussian splatting},
  author={Kim, Sieun and Lee, Kyungjin and Lee, Youngki},
  booktitle={Proceedings of the IEEE/CVF Conference on Computer Vision and Pattern Recognition},
  pages={775--783},
  year={2024}
}

@inproceedings{li2024neural,
  title={Neural video compression with feature modulation},
  author={Li, Jiahao and Li, Bin and Lu, Yan},
  booktitle={Proceedings of the IEEE/CVF Conference on Computer Vision and Pattern Recognition},
  pages={26099--26108},
  year={2024}
}

@inproceedings{li2023neural,
  title={Neural video compression with diverse contexts},
  author={Li, Jiahao and Li, Bin and Lu, Yan},
  booktitle={Proceedings of the IEEE/CVF conference on computer vision and pattern recognition},
  pages={22616--22626},
  year={2023}
}

@article{sheng2022temporal,
  title={Temporal context mining for learned video compression},
  author={Sheng, Xihua and Li, Jiahao and Li, Bin and Li, Li and Liu, Dong and Lu, Yan},
  journal={IEEE Transactions on Multimedia},
  volume={25},
  pages={7311--7322},
  year={2022},
  publisher={IEEE}
}

@inproceedings{jia2025towards,
  title={Towards practical real-time neural video compression},
  author={Jia, Zhaoyang and Li, Bin and Li, Jiahao and Xie, Wenxuan and Qi, Linfeng and Li, Houqiang and Lu, Yan},
  booktitle={Proceedings of the Computer Vision and Pattern Recognition Conference},
  pages={12543--12552},
  year={2025}
}

@article{gao2026overview,
  title={Overview and comparison of avs point cloud compression standard},
  author={Gao, Wei and Gao, Wenxu and Mu, Xingming and Peng, Changhao and Li, Ge},
  journal={arXiv preprint arXiv:2602.08613},
  year={2026}
}

@article{liu2024hemgs,
  title={Hemgs: A hybrid entropy model for 3d gaussian splatting data compression},
  author={Liu, Lei and Chen, Zhenghao and Jiang, Wei and Wang, Wei and Xu, Dong},
  journal={arXiv preprint arXiv:2411.18473},
  year={2024}
}

@article{ren2024octree,
  title={Octree-gs: Towards consistent real-time rendering with lod-structured 3d gaussians},
  author={Ren, Kerui and Jiang, Lihan and Lu, Tao and Yu, Mulin and Xu, Linning and Ni, Zhangkai and Dai, Bo},
  journal={arXiv preprint arXiv:2403.17898},
  year={2024}
}

@inproceedings{cheng2024gaussianpro,
  title={Gaussianpro: 3d gaussian splatting with progressive propagation},
  author={Cheng, Kai and Long, Xiaoxiao and Yang, Kaizhi and Yao, Yao and Yin, Wei and Ma, Yuexin and Wang, Wenping and Chen, Xuejin},
  booktitle={Forty-first International Conference on Machine Learning},
  year={2024}
}

@article{liu2024atomgs,
  title={Atomgs: Atomizing gaussian splatting for high-fidelity radiance field},
  author={Liu, Rong and Xu, Rui and Hu, Yue and Chen, Meida and Feng, Andrew},
  journal={arXiv preprint arXiv:2405.12369},
  year={2024}
}

@inproceedings{zhang2022differentiable,
  title={Differentiable point-based radiance fields for efficient view synthesis},
  author={Zhang, Qiang and Baek, Seung-Hwan and Rusinkiewicz, Szymon and Heide, Felix},
  booktitle={SIGGRAPH Asia 2022 Conference Papers},
  pages={1--12},
  year={2022}
}

\end{document}